\documentclass[11pt,letterpaper]{article}
\usepackage{inlg2017}
\usepackage{times}
\usepackage{latexsym}

\usepackage{url}
\usepackage{amsmath}
\usepackage{amssymb}

\usepackage{rotating}
\usepackage{algorithmic}
\usepackage{algorithm}

\usepackage{caption}
\usepackage{subcaption}

\usepackage[utf8x]{inputenc}

\naaclfinalcopy % Uncomment this line for the final submission
\def\naaclpaperid{***} %  Enter the naacl Paper ID here

\title{A Comparison of Neural Models for Word Ordering}

\author{Eva Hasler$^{1,2}$, Felix Stahlberg$^1$, Marcus Tomalin$^1$, Adrià de Gispert$^{1,2}$, Bill Byrne$^{1,2}$ \\\\
    $^1$Department of Engineering, University of Cambridge, UK \\
    $^2$SDL Research, Cambridge, UK\\
    {\tt \{ech57,fs439,mt126,ad465,wjb31\}@cam.ac.uk}\\
   {\tt \{ehasler,agispert,bbyrne\}@sdl.com}
}

\date{}

\begin{document}

\maketitle

\begin{abstract}
We compare several language models for the word-ordering task and propose a new {\em bag-to-sequence} neural model based on attention-based {\em sequence-to-sequence} models. We evaluate the model on a large German WMT data set where it significantly outperforms existing models. We also describe a novel search strategy for LM-based word ordering and report results on the English Penn Treebank. Our best model setup outperforms prior work both in terms of speed and quality.
\end{abstract}

\section{Introduction}
Finding the best permutation of a multi-set of words 
is a non-trivial task due to linguistic aspects such as ``syntactic structure, selective restrictions, subcategorization, and discourse considerations''~\cite{org-wordordering}. This makes the word-ordering task useful for studying and comparing different kinds of models that produce text in tasks such as general natural language generation~\cite{nlg}, image caption generation~\cite{captions}, or machine translation~\cite{bahdanau}. Since plausible word order is an essential criterion of output fluency for all of these tasks, progress on the word-ordering problem is likely to have a positive impact on these tasks as well.
Word ordering has often been addressed as {\em syntactic linearization} which is a strategy that involves using syntactic structures or part-of-speech and dependency labels ~\cite{ccg,ccg-lm,dependency-vs-ccg,zgen,transition-based-lookahead}. It has also been addressed as {\em LM-based linearization} which relies solely on language models and obtains better scores ~\cite{gyro,no-syntax}.
Recently, Schmaltz et al.~\shortcite{no-syntax} showed that recurrent neural network language models~\cite[\textsc{Rnnlm}s]{rnnlm} with long short-term memory~\cite[\textsc{Lstm}]{lstm} cells are very effective for word ordering even without any explicit syntactic information.

We continue this line of work and make the following contributions. We compare several language models on the word-ordering task and propose a {\em bag-to-sequence} neural architecture that equips an LSTM decoder with explicit context of the bag-of-words (\textsc{Bow}) to be ordered. This model
performs particularly strongly on WMT data and is complementary to an \textsc{Rnnlm}: combining both yields large BLEU gains even for small beam sizes. We also propose a novel search strategy which outperforms a previous heuristic. Both techniques together surpass prior work on the Penn Treebank at $\sim$4x the speed.

\section{Bag-to-Sequence Modeling with Attentional Neural Networks}
Given the \textsc{Bow} \{\emph{at}, \emph{bottom}, \emph{heap}, \emph{now}, \emph{of}, \emph{the}, \emph{the}, \emph{we}, \emph{'re}, \emph{.}\}, a word-ordering model may generate an output string $\mathbf{w}=$ ``\emph{now we 're at the bottom of the heap .``}. We can use an \textsc{Rnnlm}~\cite{rnnlm} to assign it a probability $P(\mathbf{w})$ by decomposing into conditionals:

\label{sec:model}
\begin{figure}
\centering
\begin{subfigure}{.23\textwidth}
  \centering
  \includegraphics[scale=0.33]{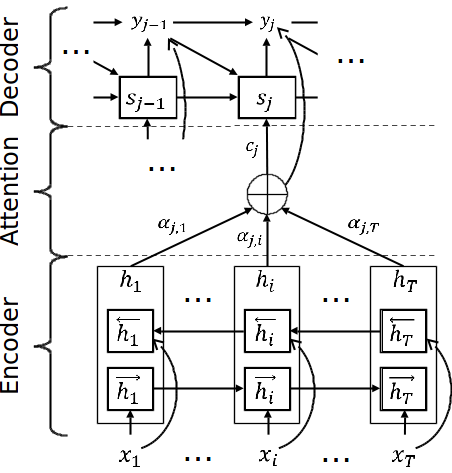}
  \caption{}
  \label{fig:seq2seq}
\end{subfigure}
\begin{subfigure}{.23\textwidth}
  \centering
  \includegraphics[scale=0.33]{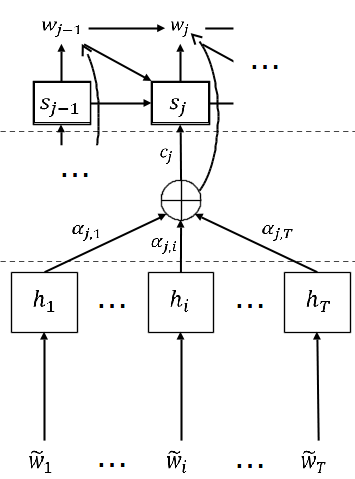}
  \caption{}
  \label{fig:bag2seq}
\end{subfigure}
\caption{(a) Attention-based {\em seq2seq} model and (b) {\em bag2seq} model used in this work.}
\label{fig:test}
\end{figure}

\begin{equation}
P(w_1^T) = \prod_{t=1}^T P(w_t|w_1^{t-1})
\label{eq:rnnlm}
\end{equation}
Since we have access to the input \textsc{Bow}s, we extend the model representation by providing the network 
additionally with the \textsc{Bow} to be ordered, thereby allowing it to focus explicitly on all tokens it generates in the output during decoding.
Thus, instead of modeling the \emph{a priori} distribution of sentences $P(\mathbf{w})$ as in Eq.~\ref{eq:rnnlm}, we condition the distribution on $\textsc{Bow}(\mathbf{w})$: 
\begin{equation}
P(w_1^T|\textsc{Bow}(\mathbf{w})) = \prod_{t=1}^T P(w_t|w_1^{t-1},\textsc{Bow}(\mathbf{w}))
\label{eq:bag2seq}
\end{equation}

This dependency is realized by the neural attention mechanism recently proposed by Bahdanau et al.~\shortcite{bahdanau}.
The resulting bag-to-sequence model ({\em bag2seq}) is inspired by the attentional sequence-to-sequence model {\sc RNNsearch} ({\em seq2seq}) proposed by Bahdanau et al.~\shortcite{bahdanau} for neural machine translation between a source sentence $\mathbf{x}=x_1^I$ and a target sentence $\mathbf{y}=y_1^J$.
Fig.~\ref{fig:seq2seq} illustrates how {\em seq2seq} generates the $j$-th target token $y_j$ using the decoder state $s_j$ and the context vector $c_j$. The context vector is the weighted sum of source side {\em annotations} $h_i$ which encode sequence information. 

To modify {\em seq2seq} for problems with unordered input, we make the encoder architecture order-invariant by replacing the recurrent layer
with non-recurrent transformations of the word embeddings, as indicated by the missing arrows between source positions in Fig.~\ref{fig:bag2seq}.
For convenience, we formalize $\textsc{Bow}(\mathbf{w})$ as sequence $\langle \tilde{w}_1, \dots, \tilde{w}_T \rangle$ in which words are sorted, e.g.\ alphabetically, so that we can refer to the $t$-th word in the \textsc{Bow}.
The model can be trained to recover word order in a sentence by using $\textsc{Bow}(\mathbf{w})=\langle \tilde{w}_1, \dots, \tilde{w}_T \rangle$ as input and the original sequence $\langle w_1, \dots, w_T \rangle$ as target.
This network architecture does not prevent words outside the \textsc{Bow} to appear in the output. Therefore, we explicitly \emph{constrain} our beam decoder by limiting its available output vocabulary to the remaining tokens in the input bag at each time step, thereby ensuring that all model outputs are \emph{valid permutations} of the input.

\section{Search}
\label{sec:search}
Beam search is a popular decoding algorithm for neural sequence models~\cite{seq2seq,bahdanau}. However, standard beam search suffers from search errors when applied to word ordering and Schmaltz et al.~\shortcite{no-syntax} reported that gains often do not saturate even with a large beam of~512. They suggested adding external unigram probabilities of the remaining words in the \textsc{Bow} as future cost estimates to the beam-search scoring function and reported large gains for an $n$-gram LM and \textsc{Rnnlm}.
We re-implement this future cost heuristic, $f(\cdot)$, and further propose a new search heuristic, $g(\cdot)$, which collects internal unigram statistics during decoding. We keep hypotheses in the beam if their score is close to a theoretical upper bound, the product of the best word probabilities given any history within the explored search space. 
For each word $\tilde{w}\in \textsc{Bow}(\mathbf{w})$ we maintain a heuristic score estimate $\hat{P}(\tilde{w})$ which we initialize to 0. Each time the search algorithm visits a new context, we update the estimates such that $\hat{P}(\tilde{w})$ is the current best score for $\tilde{w}$:
\begin{equation}
\hat{P}(\tilde{w})=\max_{c\in\mathcal{C}_t} P(\tilde{w}|c,\textsc{Bow}(\mathbf{w}))
\end{equation}
where $\mathcal{C}_t$ is the set of contexts (i.e.\ ordered prefixes in the form of $w_1^t$) explored by beam search so far. 
Thus, instead of computing a future cost, we compare the actual score of a partial hypothesis with the product of heuristic estimates of its words. 
This is especially useful for model combinations since all models are taken into account.  
We also implement hypothesis recombination to further reduce the number of search errors.
More formally, at each time step $t$ our beam search keeps the $n$ best hypotheses according to scoring function $S(\cdot)$ using partial model score $s(\cdot)$ and estimates $g(\cdot)$:

\begin{equation}
\begin{aligned}
S(w_1^t)&=& s(w_1^t)-g(w_1^t) \\
s(w_1^t)&=& \log P(w_1^t|\textsc{Bow}(\mathbf{w})) \\
g(w_1^t)&=& \sum_{w'\in w_1^t} \log \hat{P}(w')
\end{aligned}
\end{equation}

\section{Experimental Setup}
\label{sec:data}
We evaluate using data from the English-German news translation task \cite[WMT]{wmt} and using the English Penn Treebank data \cite[PTB]{ptb}.
Since additional knowledge sources are often available in practice, such as access to the source sentence in a translation scenario, we also report on bilingual experiments for the WMT task.

\subsection{Data and evaluation}
The WMT parallel training data includes {\em Europarl v7},  {\em Common Crawl}, and  {\em News Commentary v10}. 
We use {\em news-test2013} for tuning model combinations and {\em news-test2015} for testing. All monolingual models for the WMT task were trained on the German {\em news2015} corpus ($\sim$51.3M sentences). For PTB, we use preprocessed data by Schmaltz et al.~\shortcite{no-syntax} for a fair comparison ($\sim$40k sentences for training). 
We evaluate using the multi-bleu.perl script for WMT and mteval-v13.pl for PTB. 

\subsection{Model settings}
For WMT, the {\em bag2seq} parameter settings follow the recent NMT systems trained on WMT data. We use a 50k vocabulary,  620 dimensional word embeddings and 1000 hidden units in the decoder \textsc{Lstm} cells. On the encoder side, the input tokens are embedded to form annotations of the same size as the hidden units in the decoder. 
The \textsc{Rnnlm} is based on the ``large'' setup of Zaremba et al.~\shortcite{rnnlm-google} which uses an \textsc{Lstm}. 
\textsc{Nplm}, a  5-gram neural feedforward language model, was trained for 10 epochs with a vocabulary size of 100k, 150 input and output units, 750 hidden units and 100 noise samples \cite{nplm}. The $n$-gram language model is a 5-gram model estimated with \textsc{Srilm}~\cite{ngram}. 
For the bilingual setting, we implemented a {\em seq2seq} NMT system following Bahdanau et al.~\shortcite{bahdanau} 
using a beam size of 12 in line with recent NMT systems for WMT \cite{wmt2016}. \textsc{Rnnlm}, {\em bag2seq} and {\em seq2seq} were implemented using TensorFlow~\cite{tensorflow}
\footnote{\url{https://github.com/ehasler/tensorflow}} and we used {\em sgnmt} for beam decoding\footnote{\url{https://github.com/ucam-smt/sgnmt}}.

Following Schmaltz et al.~\shortcite{no-syntax}, our neural models for PTB have a vocabulary of 16,161 incl. two different $unk$ tokens and the \textsc{Rnnlm} is based on the ``medium'' setup of Zaremba et al.~\shortcite{rnnlm-google}. \emph{bag2seq}  
uses 300 dimensional word embeddings and 500 hidden units in the decoder \textsc{Lstm}. We also compare to \textsc{Gyro} \cite{gyro} which explicitly targets the word-ordering problem. We extracted 1-gram to 5-gram phrase rules from the PTB training data and used an $n$-gram LM 
for decoding.
For model combinations, we combine the predictive distributions in a log-linear model and tune the weights by optimizing BLEU on the validation set with the BOBYQA algorithm~\cite{bobyqa}.

\setlength{\tabcolsep}{2pt}
% With hypo recombination and f(\cdot) for single models
% With bow_equiv_vocab for models/model combinations not containing srilm
\begin{table}[t!]
\begin{center}
\begin{tabular}{ccccc|c}
{\bf \textsc{Rnnlm}} & {\bf \textsc{Nplm}} & {\bf $n$-gram} & {\bf bag2seq} & {\bf seq2seq} & {\bf \textsc{Bleu}} \\
\hline
$\checkmark$ & & & & & 29.4 \\
& $\checkmark$ & & & & 30.3 \\
& & $\checkmark$ & & & 32.5 \\ 
& & & $\checkmark$ & & \bf 33.6 \\
$\checkmark$ & \checkmark & \checkmark & & & 34.9 \\
$\checkmark$ & \checkmark & \checkmark & $\checkmark$ & & \bf 39.4 \\ 
\hline
& & & & $\checkmark$ & 49.7 \\
& & & $\checkmark$ & $\checkmark$ & \bf 52.6 \\
$\checkmark$ & \checkmark & \checkmark & & \checkmark & 51.3 \\
$\checkmark$ & \checkmark & \checkmark & $\checkmark$ & \checkmark & \bf 53.1 \\
\end{tabular}
\end{center}
\caption{\label{tab:wmt} German word ordering on {\em news-test2015} with {\em beam=12}, single models/combinations. Monolingual models use  heuristic $f(\cdot)$, {\em bag2seq} as a single model and bilingual models use no heuristic.}
\end{table}

\section{Results}

\subsection{Word Ordering on WMT data}
The top of Tab.~\ref{tab:wmt} shows that {\em bag2seq} outperforms all other language models by up to 4.2 BLEU on ordering German (bold numbers highlight its improvements). 
This suggests that explicitly presenting all available tokens to the decoder during search enables it to make better word order choices.
A combination of \textsc{Rnnlm}, \textsc{Nplm} and $n$-gram LM yields a higher score than the individual models, but further adding {\em bag2seq} yields a large gain of 4.5 BLEU confirming its suitability for the word-ordering task.

In the bilingual setting in the bottom of Tab.~\ref{tab:wmt}, the {\em seq2seq} model is given English input text and the beam decoder is constrained to generate permutations of German \textsc{Bow}s. This is effectively a translation task with knowledge of the target \textsc{Bow}s and {\em seq2seq} provides a strong baseline since it uses source sequence information. 
Still, adding {\em bag2seq} yields a 2.9 BLEU gain and adding it to the combination of all other models still improves by 1.8 BLEU. This suggests that it could also help for machine translation rescoring by selecting hypotheses that constitute good word orderings.

\subsection{Word Ordering on the Penn Treebank}
\label{ptb}
Tab.~\ref{tab:search} shows the performance of different models and search heuristics on the Penn Treebank: using no heuristic ({\em none}) vs. $f(\cdot)$ and $g(\cdot)$ described in Section~\ref{sec:search}. Numbers in bold mark the best result for a given model.
We compare against the LM-based method of de Gispert et al.~\shortcite{gyro} and the $n$-gram and \textsc{Rnnlm} (\textsc{Lstm}) models of Schmaltz et al.~\shortcite{no-syntax}, of which the latter achieves the best BLEU score of 42.7. 
We can reproduce or surpass prior work for $n$-gram and \textsc{Rnnlm} and show that $g(\cdot)$ outperforms $f(\cdot)$ for these models. This also holds when adding a 900k sample from the English Gigaword corpus as proposed by Schmaltz et al.~\shortcite{no-syntax}.\footnote{Results omitted from Tab.~\ref{tab:search} to save space.} 
However, {\em bag2seq} underperforms \textsc{Rnnlm} at this large beam size.

\setlength{\tabcolsep}{4pt}
\begin{table}[t!]
\begin{center}
\begin{tabular}{l|ccc}
\bf Model & {\em none} & $f(\cdot)$ & $g(\cdot)$ \\\hline
\multicolumn{1}{l}{\em Previous work} & \multicolumn{3}{c}{\em beam=512} \\\hline
\textsc{Gyro}\footnotemark & 42.2 & -- & -- \\
\textsc{Ngram-512} &  -- & 38.6 & -- \\
\textsc{Lstm-512} & -- & 42.7 & -- \\\hline
\multicolumn{1}{l}{\em This work} & \multicolumn{3}{c}{\em beam=512}\\\hline
$n$-gram & 35.7 & 38.6 & \bf38.9 \\
\textsc{Rnnlm} & 38.6 & 43.2 & \bf44.2 \\
{\em bag2seq} & \bf 37.1 & 33.6 & \bf 37.1 \\
\end{tabular}
\end{center}
\caption{\label{tab:search} BLEU scores for PTB word-ordering task (test).  
\textsc{Ngram-512} and \textsc{Lstm-512} are quoted from Schmaltz et al. (2016).}
\end{table}
\footnotetext{Note that this model has an advantage because longer sentences are processed in chunks of maximum length 20.}

\begin{figure}
\centering

   \includegraphics[trim=45 200 0 280, clip, scale=0.45]{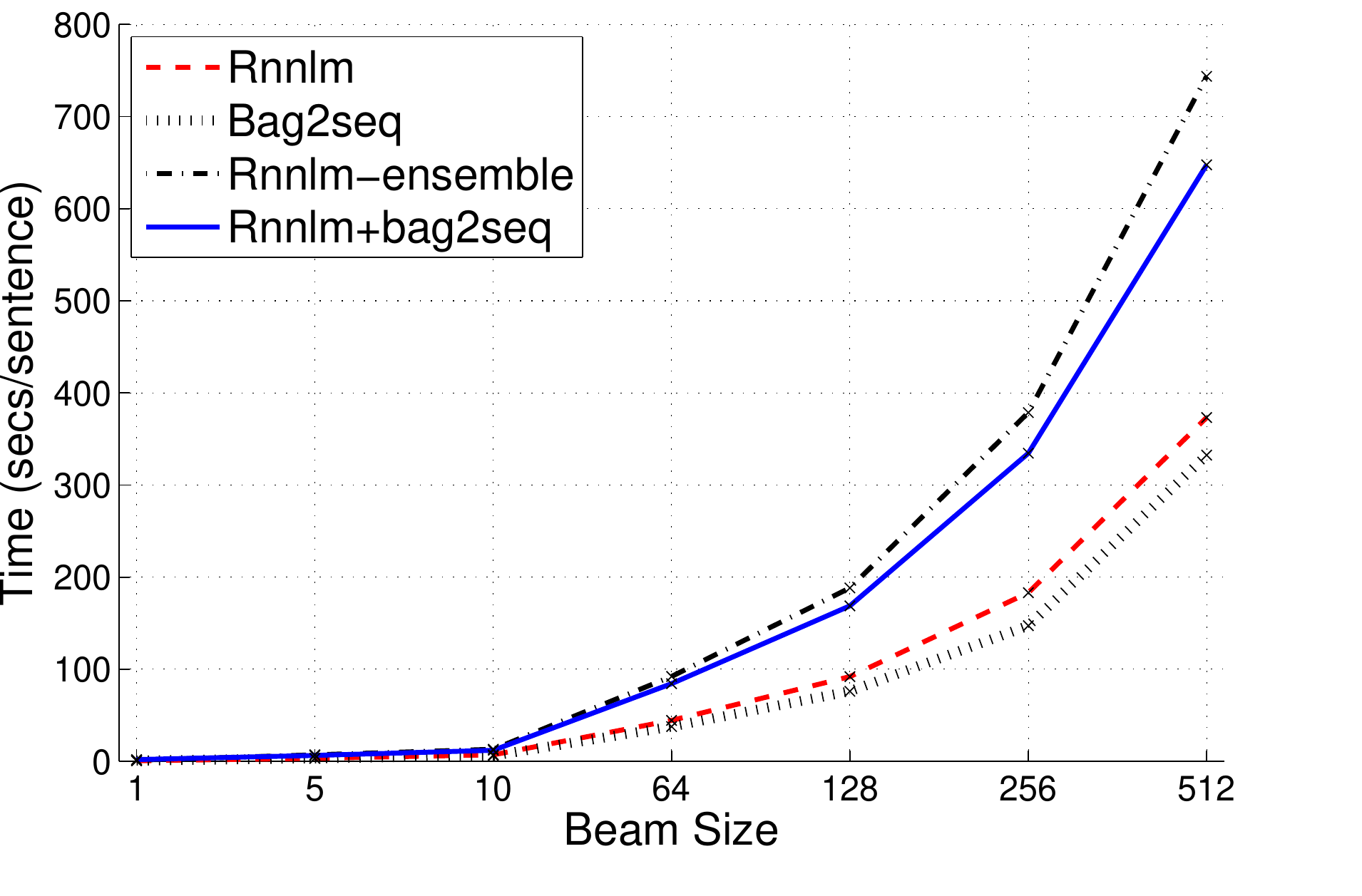}
\caption{Decoding time in relation to beam size for PTB word ordering task (test).}
\label{fig:time}
\end{figure}

Since decoding is slow for large beam sizes, we compare {\em bag2seq} to the $n$-gram and \textsc{Rnnlm} using a small beam of size 5 in Tab.~\ref{tab:ptb}.
The first three rows show that decoding without heuristics is much easier with {\em bag2seq} and outperforms $n$-gram and \textsc{Rnnlm} by a large margin with 33.4 BLEU. The \textsc{Rnnlm} needs heuristic $f(\cdot)$ to match this performance.
For {\em bag2seq}, using heuristic estimates is worse than just using its partial scores for search. We suspect that its partial model scores are obfuscated by the heuristic estimates and the amount of their contribution should probably be tuned on a heldout set. 
Using the same beam size, ensembles yield better results but the best results are achieved by combining \textsc{Rnnlm} and {\em bag2seq} (37.9 BLEU). This confirms our findings on WMT data that these models are highly complementary for word ordering.
The results for beam=64 follow this pattern and identify an interaction between heuristics and beam size. While we get the best results for beam=5 using $f(\cdot)$, heuristic $g(\cdot)$ seems to perform better for larger beams, perhaps because the internal unigram statistics become more reliable.
Finally, \textsc{Rnnlm}+{\em bag2seq} with $g(\cdot)$ and beam=64 outperforms \textsc{Lstm-512} by 0.8 BLEU. This is significant because decoding in this configuration is also $\sim$4x faster than decoding with a single \textsc{Rnnlm} and beam=512  
as shown in Fig.~\ref{fig:time}.

\setlength{\tabcolsep}{4pt}
\begin{table}[t!]
\begin{center}
\begin{tabular}{l|ccc}
{\bf Model} & {\em none} & $f(\cdot)$ & $g(\cdot)$ \\\hline
\multicolumn{1}{l}{} & \multicolumn{3}{c}{\em beam=5} \\\hline
$n$-gram & 23.3 & \bf 30.1 & 26.5 \\
\textsc{Rnnlm} & 24.5 & \bf 33.6 & 29.7  \\
{\em bag2seq} & \bf 33.4 & 27.0 & 31.7 \\\hline
\textsc{Rnnlm}-ensemble & 25.5 & \bf 34.2 & 30.6 \\
{\em bag2seq}-ensemble & 34.8 & \bf 35.1 & 32.8 \\
\textsc{Rnnlm}+{\em bag2seq} & 35.7 & \bf 37.9 & 34.4 \\\hline
\multicolumn{1}{l}{} & \multicolumn{3}{c}{\em beam=64} \\\hline
\textsc{Rnnlm} & 34.6 & 40.9 & \bf 42.5 \\
{\em bag2seq} & 36.2 & 31.4 & \bf 36.5 \\\hline
\textsc{Rnnlm}-ensemble & 35.4 & 42.4 & \bf 43.2 \\
\textsc{Rnnlm}+{\em bag2seq} & 40.5 & 43.1 & \bf 43.5 \\
\hline
\end{tabular}
\end{center}
\caption{\label{tab:ptb}BLEU scores for PTB word-ordering task for different search heuristics and beam sizes (test).}
\end{table}

\section{Conclusion}
We have compared various models for the word-ordering task and proposed a new model architecture inspired by attention-based sequence-to-sequence models that helps performance for both German and English tasks. We have also proposed a novel search heuristic and found that using a model combination together with this heuristic and a modest beam size provides a good trade-off between speed and quality and outperforms prior work on the PTB task.

\section*{Acknowledgments}
This work  was partially supported  by the U.K.\ Engineering and Physical Sciences Research Council (EPSRC grant EP/L027623/1).
We thank the authors of Schmaltz et al.~\shortcite{no-syntax} for sharing their preprocessed data and helping to reproduce their results.

\bibliography{inlg2017}
\bibliographystyle{inlg2017}

\end{document}